# Selective Prediction and Uncertainty-Aware Referral for Pap Smear Classification

Nisreen Albzour*, Sarah S. Lam

School of Systems Science and Industrial Engineering, Binghamton University, Binghamton, NY 13902, USA

* Corresponding author: nalbzour@binghamton.edu

ORCID: https://orcid.org/0009-0000-9317-340X

## Abstract

Deep learning models for cervical cytology are almost always evaluated as if every prediction must be acted upon, yet a screening system deployed alongside a cytopathologist need not classify every slide: it can defer the cases it is least certain about. Evaluating such a system requires asking not only how often it is correct, but whether its confidence ranks its errors to the bottom. This paper studies selective prediction and uncertainty-aware referral on the Herlev Pap smear dataset under a binary Normal-versus-Abnormal formulation. Two lightweight transformer backbones (Swin-Tiny, TinyViT-5M) are fine-tuned on Herlev from ImageNet-pretrained weights with weighted random sampling, calibrated by post-hoc temperature scaling fit on a held-out calibration subset, and compared against a soft-voting ensemble of both models. Discrimination is reported alongside expected calibration error (ECE) and, as the primary endpoint, the area under the risk-coverage curve (AURC). No statistically significant difference was detected between the two configurations in accuracy or macro-F1, yet the ensemble halves AURC (0.0022 vs. 0.0045, a 51.8% reduction, lower in all five folds) and extends the coverage at which zero errors are made from 18.3% to 72.8% of the pooled test predictions. The same ensemble is nonetheless worse calibrated in absolute terms (ECE 0.0339 vs. 0.0247) and produces more false negatives (14 vs. 10). These results separate two properties that are frequently conflated: the ability to rank predictions by trustworthiness, and the accuracy of the confidence values themselves. Ensembling improves the former while degrading the latter, and the former directly governs the observed risk-coverage tradeoff, whereas the latter governs the interpretation of the reported confidence values.

## 1. Introduction

Cervical cancer remains a substantial global health burden, and its incidence and mortality remain well above the threshold set by the WHO elimination initiative in most countries, with marked inequalities across levels of human development [1]. Screening by Pap smear is the principal tool for early detection, but manual examination is labor-intensive, requires expert cytologists, and is subject to delays and inconsistency between readers [2]. Deep learning has accordingly been applied to cervical cell classification, with convolutional, transformer, and ensemble architectures all reporting high discrimination on public benchmarks [2]. The task itself, however, is one of triage performed under a heavy and uneven workload: the overwhelming majority of slides examined in a screening program are negative, and the cytopathologist's time is most valuable on the small fraction that are not. An automated classifier deployed in this setting is therefore not required to render a verdict on every slide. It may instead abstain on the cases it finds ambiguous, referring them for human review while auto-clearing the remainder. This is the setting of selective prediction, and the quantity that governs it is not accuracy but the relationship between a model's confidence and its errors.

The distinction matters because two models with identical accuracy can behave very differently under referral. If one model's mistakes are concentrated among its least confident predictions, a low confidence threshold will remove nearly all of them while retaining most of the workload. If another model's mistakes are scattered across the confidence range, no threshold removes them without also discarding a large volume of correct predictions. The risk-coverage curve makes this behavior explicit by plotting the error rate among retained predictions against the fraction retained, and the area under that curve (AURC) summarizes it in a single number [3], [4].

AURC is not, however, a calibration metric, and this paper argues that the difference is easy to lose sight of. Expected calibration error asks whether a stated confidence of 0.9 corresponds to being correct 90% of the time. AURC asks only whether predictions the model is more confident about are more often correct than those it is less confident about. The first is a statement about absolute probability values; the second is a statement about their ordering. A model can rank its predictions perfectly while systematically overstating or understating its confidence, and post-hoc temperature scaling, which rescales confidence without reordering predictions [5], changes the second not at all while changing the first substantially.

Uncertainty estimation and abstention have been studied in medical imaging: rejecting low-confidence samples on the basis of predictive uncertainty has been shown to improve the reliability of the predictions that are retained [6], and selective prediction has recently been extended from classification to segmentation, where its authors note that selective classification has received considerable attention while only a few studies have explored the segmentation setting [7]. A related principle argues for reporting reliability at the level of the individual case rather than only in aggregate: converting a continuous quality score into discrete per-case risk categories makes the reliability of each prediction explicit where aggregate metrics alone would obscure it [8]. In cervical cytology specifically, evaluation continues to center on accuracy, F1-score, sensitivity, specificity, and the area under the ROC curve (AUROC). Calibration has received limited attention on this task, few studies have evaluated referral behavior, and the joint effect of ensembling on calibration error and on confidence-based error ranking remains underexplored on Herlev.

This paper reports a case in which the two properties diverge cleanly. On the Herlev Pap smear dataset, a soft-voting ensemble of two lightweight transformer backbones shows no statistically significant difference from the better of its two members in accuracy or macro-F1, and is measurably worse calibrated, yet it halves AURC and quadruples the coverage at which no errors are made. The contribution is threefold. First, the study is conducted entirely on Herlev, with all models trained directly on that dataset rather than transferred from another source, and with temperature scaling fit on a calibration subset carved out of each cross-validation fold's training portion so that no sample used to fit a temperature is ever used to evaluate it. Second, selective prediction is evaluated as a primary endpoint rather than as an afterthought, with AURC reported per fold so that the consistency of the effect can be assessed rather than only its pooled magnitude. Third, the calibration and referral results are reported together, which is what makes the divergence between them visible.

## 2. Related Work

This section reviews the three bodies of work on which the study draws: deep learning for Pap smear classification on the Herlev dataset (Section 2.1), confidence calibration in medical image classification (Section 2.2), and selective prediction with a reject option (Section 2.3). The first is mature but narrow in what it measures; the second is well developed and largely disjoint from the first; the third has grown up mostly outside medical imaging. This paper sits at their intersection.

### 2.1 Pap Smear Classification on Herlev

The Herlev dataset comprises 917 single-cell Pap smear images annotated into seven native subtypes, three normal and four abnormal, and remains among the most widely used benchmarks in cervical cytology [9]. Because the seven-class problem is difficult at this data scale, most published work collapses the labels into a binary Normal-versus-Abnormal task, and this paper follows that convention.

The methodological arc of this literature is by now familiar. Zhang et al. established the deep-learning baseline with DeepPap, a transfer-learned convolutional network operating on nucleus-centered image patches with test-time score aggregation, reporting 98.3% accuracy, an AUC of 0.99, and 98.3% specificity on Herlev under five-fold cross-validation [10]. Subsequent work has largely refined the input representation or the architecture rather than the evaluation. Lin et al. asked whether cell morphology could be modeled directly by a convolutional network alongside image appearance, comparing AlexNet, GoogLeNet, ResNet, and DenseNet backbones across the two-, four-, and seven-class formulations [11]. Jain et al. paired four pretrained convolutional backbones with metaheuristic hyperparameter optimization [12], while Kaur et al. benchmarked sixteen transfer-learning architectures spanning the VGG, ResNet, DenseNet, MobileNet, Xception, and Inception families across both Herlev and SIPaKMeD [13]. Our own earlier work on Herlev spans this same arc: a U-Net segmentation stage coupled to a classifier, which found that segmenting cells before classification improved precision and F1 only marginally [14]; an evaluation of vision transformers with attention-based interpretability under the same binary formulation [15]; and a benchmark comparing ViT-Tiny against convolutional baselines on cervical cell images with attention to deployment cost [16]. None of the three reports calibration, and none permits the model to abstain.

What unites these studies is not the architecture but the evaluation. Accuracy, F1-score, sensitivity, specificity, and AUROC are reported almost universally; calibration is reported almost never, and abstention or referral not at all. The consequence is that a decade of architectural progress on Herlev has optimized and measured a single property, the correctness of the top-1 prediction, while leaving untouched the question of whether the confidence attached to that prediction can be trusted, or used. Where ensembles appear in this literature, they serve as accuracy-boosting devices, evaluated on the same discrimination metrics as their members. Across this body of work, models are evaluated under the implicit assumption that every prediction is acted upon.

### 2.2 Calibration in Medical Image Classification

Modern neural networks are frequently poorly calibrated even when highly accurate. Temperature scaling is the simplest effective correction: a single scalar divides the logits before the softmax, rescaling confidence without altering the ranking of predicted classes [5]. Its appeal in clinical settings is that it is post-hoc and requires only a held-out calibration set. For an individual classifier, temperature scaling preserves class ordering and argmax predictions; however, independently calibrating ensemble members before probability averaging may alter the ensemble output.

Within medical imaging the literature has developed along two lines. One pursues calibration as a training objective. Liang et al., observing that work in this domain had focused on classification accuracy while ignoring uncertainty quantification, introduced an auxiliary differentiable term that penalizes a model when training loss falls without a corresponding gain in accuracy, reducing expected calibration error during optimization rather than after it [17]. The other line treats calibration as a post-hoc correction and asks which correction is best: Carse et al. compared temperature scaling against training-time alternatives including focal loss and label smoothing on dermatology and histopathology datasets [18]. Robustness to imperfect supervision has also received attention. Penso et al. incorporate estimated label-noise structure into the calibration procedure [19], a concern that is recognized in cytology, where diagnostic agreement between observers is imperfect. Our own recent study on liquid-based cervical cytology found that temperature scaling reduced expected calibration error by 85 to 95 percent across every architecture and ensemble size tested, without any change to discrimination [20].

Expected calibration error is the near-universal endpoint of this work, and it measures a specific thing: whether a stated confidence of p corresponds to being correct a fraction p of the time. It is a statement about the numerical accuracy of probability values, evaluated by binning predictions by confidence and comparing mean confidence to empirical accuracy within each bin. What ECE does not measure is whether predictions the model is more confident about are more often correct than those it is less confident about. A model whose confidences are uniformly inflated by a constant factor is badly calibrated but may rank its predictions perfectly; conversely, a model whose average confidence matches its average accuracy may scatter its errors across the confidence range. The two properties are logically independent, and to our knowledge no study in this literature evaluates both on cervical cytology.

### 2.3 Selective Prediction and the Reject Option

The idea that a classifier should be permitted to abstain dates to Chow's formulation of the optimum error-reject tradeoff [21]. El-Yaniv and Wiener placed the modern treatment on formal footing, defining the risk-coverage tradeoff that characterizes any selective classifier [3], and Geifman and El-Yaniv extended it to deep networks, first through post-hoc selection with guaranteed risk control [22] and subsequently through SelectiveNet, which learns the selection function jointly with the classifier rather than thresholding the confidence of a pre-trained network [23]. The selection function need not be learned: the maximum softmax probability is a strong and widely used baseline [24], and it is the confidence score adopted in this paper.

Evaluating a selective classifier requires summarizing an entire curve rather than a point. The risk-coverage curve plots the error rate among retained predictions against the fraction retained, and the area under it (AURC) provides a threshold-free scalar summary. Zhou et al. give a formal statistical characterization of the population AURC together with Monte Carlo plug-in estimators, observing that AURC computed empirically from a given dataset is susceptible to bias and variance in the finite-sample regime rather than the underlying population [4], a caution directly relevant to a dataset of 917 images, and one reason this paper reports AURC per fold rather than only pooled. Pugnana and Ruggieri make the underlying distinction explicit from the opposite direction, observing that selective classification whose performance is measured by a distributive loss such as error rate is not the same problem as selective classification measured by a ranking metric, and that methods succeeding at the former can fail at the latter [25]. This is the same separation that motivates the present study: ECE measures the accuracy of confidence values, whereas AURC measures how well those values rank correct predictions above incorrect ones.

Deep ensembles are the standard non-Bayesian approach to predictive uncertainty [26], and uncertainty-aware ensembles have been applied to Pap smear classification specifically [27]. Such work typically reports accuracy, and sometimes calibration, but rarely referral behavior. The two evaluation axes are usually pursued in isolation: the calibration literature surveyed in Section 2.2 reports ECE without risk-coverage analysis, while the selective-prediction literature reports AURC without asking whether the underlying confidences are numerically accurate.

Few studies examine the two together, and their joint behavior under ensembling is not obviously benign. Averaging the predicted probabilities of several models pulls confidence values toward the interior of the range, which may worsen ECE, while simultaneously suppressing the idiosyncratic high-confidence errors of any single member, which may improve AURC. Whether such an advantage reflects a genuine improvement rather than an artifact is what external, cross-cohort validation is meant to establish, and the same setting shows that distinct reliability properties need not move together, as seen when the gains of hybrid deep-boosted ensembles for breast cancer survival prediction were re-examined across cohorts [28], [29]. To our knowledge, the joint effect of ensembling on expected calibration error and on the area under the risk-coverage curve remains underexplored on Herlev, and no prior work on this dataset reports both alongside referral behavior. The present study measures both, on the same folds and the same held-out predictions, and finds that they move in opposite directions.

## 3. Methodology

All experiments were conducted directly on the Herlev dataset. No models or checkpoints were transferred from any other cervical cytology dataset; every model reported here was fine-tuned from ImageNet-pretrained weights on Herlev alone.

Figure 1 summarizes the full experimental pipeline; the remainder of this section describes each stage in turn.

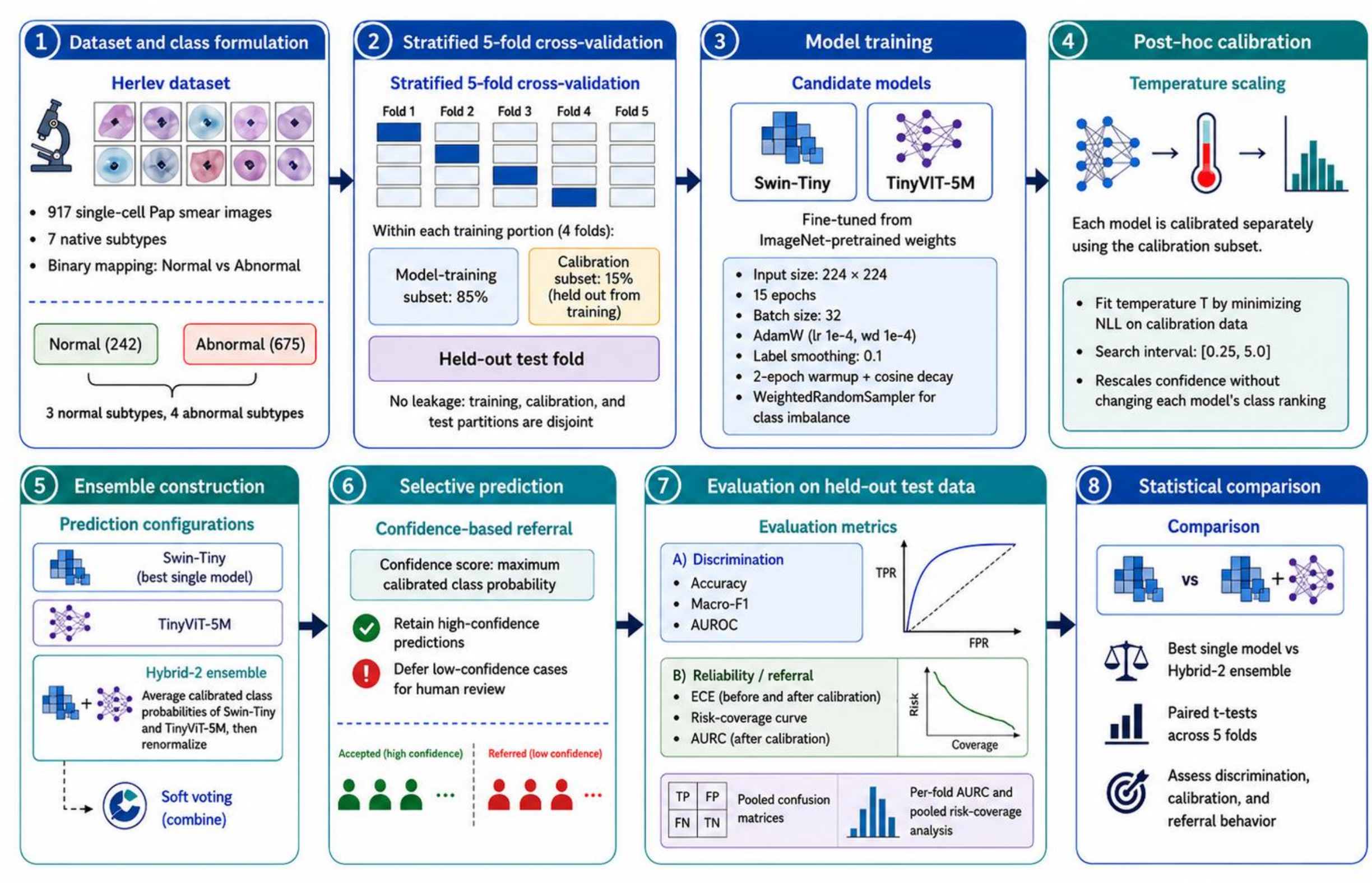

***Figure 1. Overview of the experimental pipeline: dataset and binary class formulation, the leakage-safe stratified five-fold cross-validation design, training of the two candidate models, post-hoc temperature scaling on the held-out calibration subset, construction of the Hybrid-2 soft-voting ensemble, confidence-based selective prediction, evaluation on held-out test data, and the statistical comparison of the best single model against the ensemble.***

## 3.1 Dataset

Herlev comprises 917 single-cell Pap smear images distributed across seven native subtypes. Following standard practice, the three normal subtypes (superficial, intermediate, and columnar) are mapped to a Normal class and the four abnormal subtypes (light, moderate, and severe dysplasia, and carcinoma in situ) to an Abnormal class. The resulting binary problem is skewed toward the abnormal category, which is the reverse of the skew found in a screening population and is an artifact of how the dataset was assembled. Table 1 reports the class distribution and the corresponding inverse-frequency sampling weights used to counteract this imbalance during training (Section 3.4). Table 2 reports the underlying seven-subtype composition. Figure 2 shows one representative image from each of the seven native subtypes.

***Table 1. Class distribution and sampling weights.***

| Class | Count | Sampling Weight |
|---|---|---|
| Normal | 242 | 0.004132 |
| Abnormal | 675 | 0.001481 |

*Table 2. Native seven-subtype composition of the Herlev dataset, and the binary class to which each subtype is mapped.*

| Subtype | Count | Mapped Class |
|---|---|---|
| Severe dysplastic | 197 | Abnormal |
| Light dysplastic | 182 | Abnormal |
| Carcinoma in situ | 150 | Abnormal |
| Moderate dysplastic | 146 | Abnormal |
| Normal columnar | 98 | Normal |
| Normal superficial | 74 | Normal |
| Normal intermediate | 70 | Normal |

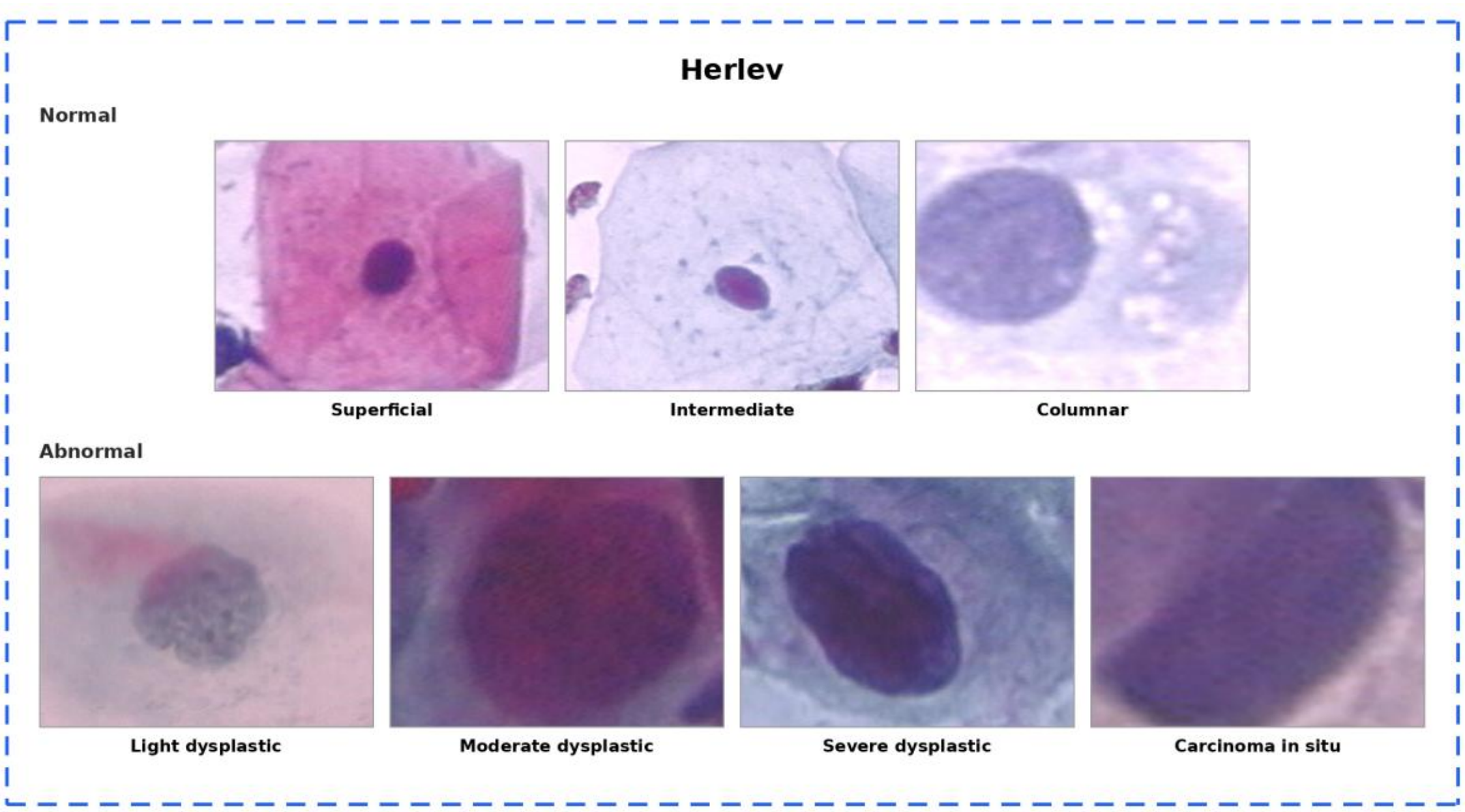

*Figure 2. Representative Herlev images for each of the seven native subtypes, grouped by the binary class to which they are mapped. Each panel shows an equally sized window onto one cell: images are scaled to fill the panel and the overflow is cropped, so aspect ratios are preserved and no cell is stretched. The panels are therefore not to a common scale, and the apparent sizes here do not reflect the true relative sizes of the cells, which differ by more than an order of magnitude in pixel area. Images are upscaled for display.*

### 3.2 Class Formulation

The binary Normal-versus-Abnormal formulation is used throughout. This is a deliberate simplification relative to the seven-subtype labels, adopted because it is the convention in the Herlev literature and because the referral question this paper studies, whether a slide can be auto-cleared or must be reviewed, is itself binary. The cost of the simplification is that it discards the distinction between low-grade and high-grade abnormality, which a clinical triage system would need.

### 3.3 Candidate Models and Ensemble

Two lightweight transformer architectures were selected as the candidate pool: Swin-Tiny and TinyViT-5M. The choice of compact transformer backbones follows evidence that such models remain competitive with heavier convolutional baselines at reduced deployment cost on cervical cell images [16]. Each was fine-tuned directly on Herlev from ImageNet-pretrained weights. A single soft-voting ensemble, denoted Hybrid-2, was constructed by averaging the predicted class probabilities of both models and renormalizing. Because the candidate pool contains exactly two models, Hybrid-2 comprises the entire pool; no model selection is performed, and the comparison reported in this paper is therefore between the better of the two individual models and the average of both, not between a selected subset and a larger pool.

### 3.4 Class-Imbalance Handling

Class imbalance was addressed during training using a WeightedRandomSampler with inverse class-frequency sampling weights, so that images from the minority Normal class are sampled more frequently per epoch than their raw frequency in the dataset. This changes each sample's sampling probability during training; it does not alter the underlying dataset or its true class distribution, which is reported separately in Table 1 alongside the sampling weights it implies.

### 3.5 Calibration Protocol

Post-hoc temperature scaling was applied to calibrate each model's predicted probabilities. To avoid the optimistic calibration estimates that result when the same data is used both to fit a temperature and to evaluate it, each cross-validation fold's training portion was further split into a model-training subset (85%) and a calibration subset (15%, stratified by class), with the calibration subset held out from model training entirely. The temperature was fit by minimizing negative log-likelihood on the calibration subset only, over the bounded interval [0.25, 5.0]. Final metrics, both before and after calibration, are computed exclusively on the fold's held-out test partition, which is never used for either training or calibration.

As a specific leakage check: within every cross-validation fold, the temperature-scaling parameter is fit exclusively on the calibration subset described above, and every accuracy, macro-F1, AUROC, ECE, and AURC value reported in this paper is computed exclusively on that fold's held-out test partition. No sample used to fit a temperature is ever also used to evaluate it, and no sample used for either training or calibration is ever included in a reported test metric. This partitioning discipline follows the leakage-aware benchmarking protocol we developed to keep overlapping partitions from inflating reported reliability [30], [31].

### 3.6 Selective Prediction and the Risk-Coverage Curve

Selective prediction is evaluated using the maximum calibrated class probability as the confidence score, which is the standard baseline selection function [24]. For a coverage level c, the c fraction of test predictions with the highest confidence is retained and the remainder is deferred; the risk at that coverage is the error rate among the retained predictions. Sweeping c from 1/n to 1 traces the risk-coverage curve, where n is the number of test predictions.

The area under this curve (AURC) is computed by trapezoidal integration over the swept coverage grid and normalized by the width of that grid, which spans from 1/n to 1 rather than from 0 to 1. This normalization differs from definitions that integrate over the full unit interval, and yields slightly larger values than an unnormalized integral would; it is stated here explicitly so that the absolute magnitudes reported below are interpreted correctly. All comparisons in this paper are made between configurations evaluated under the identical normalization, so the relative differences are unaffected. AURC is computed on after-calibration probabilities throughout.

AURC is reported per fold, so that a mean and standard deviation across the five folds can be given and the consistency of any difference assessed. A pooled risk-coverage curve, obtained by concatenating the held-out test predictions of all five folds, is also produced for visualization. The pooled curve and the per-fold mean are not the same quantity: pooling mixes predictions whose confidence values were produced under different fold-specific temperatures, so the pooled AURC can differ appreciably from the mean of the per-fold AURCs. Both are reported below, and the per-fold statistics are used for all inferential claims.

### 3.7 Cross-Validation and Evaluation Metrics

Stratified five-fold cross-validation was used, with training, calibration, and test partitions constructed as described above within each fold. Four metrics were computed on the held-out test partition of every fold: accuracy, macro-averaged F1, AUROC, and expected calibration error (ECE, 15 bins). Each is reported both before calibration (raw softmax outputs) and after calibration (temperature-scaled outputs). AURC is reported on calibrated outputs only. Models were trained for 15 epochs with batch size 32 using AdamW (learning rate 1e-4, weight decay 1e-4), cross-entropy loss with label smoothing 0.1, a two-epoch linear warmup followed by cosine decay, and 224x224 inputs.

## 4. Results

Table 3 reports discrimination metrics for the best individual model (Swin-Tiny) and the two-model ensemble (Hybrid-2), averaged across the five cross-validation folds. Swin-Tiny was the better of the two individual models under the after-calibration ranking criterion and is reported as the best single model throughout.

***Table 3. Discrimination metrics for the best single model (Swin-Tiny) and the Hybrid-2 ensemble, before and after temperature scaling, averaged across 5 cross-validation folds. All metrics computed on held-out test partitions only.***

| Config | Acc. (before) | Acc. (after) | Macro-F1 (before) | Macro-F1 (after) | AUROC (before) | AUROC (after) |
|---|---|---|---|---|---|---|
| Swin-Tiny | 0.9662 | 0.9662 | 0.9559 | 0.9559 | 0.9907 | 0.9907 |
| Hybrid-2 | 0.9706 | 0.9727 | 0.9621 | 0.9650 | 0.9953 | 0.9947 |

Accuracy, macro-F1, and AUROC are unchanged to four decimal places before and after calibration for Swin-Tiny, as expected: temperature scaling divides all logits by a single positive scalar and therefore cannot reorder predicted classes. The small changes visible for Hybrid-2 arise because calibration is applied to each member independently before their probabilities are averaged, so the averaging operation combines differently scaled distributions and can alter the ensemble's argmax.

The two configurations are close on the metrics most commonly reported. Across the five folds a paired t-test finds no significant difference in accuracy (0.9727 vs. 0.9662, $p = 0.109$) or macro-F1 (0.9650 vs. 0.9559, $p = 0.094$). The ensemble does achieve a significantly higher AUROC (0.9947 vs. 0.9907, $p = 0.014$). On the basis of Table 3 alone, a reader would reasonably conclude that ensembling buys a marginal and largely non-significant improvement.

***Table 4. Reliability and selective-prediction metrics for the same two configurations. ECE is reported before and after calibration; AURC is computed on calibrated outputs and reported as the mean and standard deviation across the five folds, alongside the pooled value.***

| Config | ECE (before) | ECE (after) | AURC (mean) | AURC (std) | AURC (pooled) |
|---|---|---|---|---|---|
| Swin-Tiny | 0.0527 | 0.0247 | 0.0045 | 0.0018 | 0.0072 |
| Hybrid-2 | 0.0653 | 0.0339 | 0.0022 | 0.0008 | 0.0020 |

Table 4 tells a different story. Temperature scaling substantially improves calibration for both configurations, reducing ECE by 53.0% for Swin-Tiny and 48.0% for Hybrid-2. But after calibration the ensemble is less well calibrated than the best single model, with a mean ECE of 0.0339 against Swin-Tiny's 0.0247, a gap that holds in four of the five folds. Its advantage lies entirely in the selective-prediction column: AURC falls from 0.0045 to 0.0022, a 51.8% reduction, and this difference is significant under a paired t-test across folds ($p = 0.028$) and consistent in direction in all five folds. Because this comparison is based on only five overlapping cross-validation folds, the inferential result should be interpreted cautiously.

Figure 3 shows the per-fold AURC values that underlie this comparison. The ensemble achieves the lower AURC in every fold, and its fold-to-fold variability is less than half that of the single model (standard deviation 0.0008 vs. 0.0018). The consistency, rather than the magnitude of the pooled difference, is the stronger evidence here: with only five folds, a difference of this size that reverses in no fold is more informative than a large mean gap driven by one or two folds.

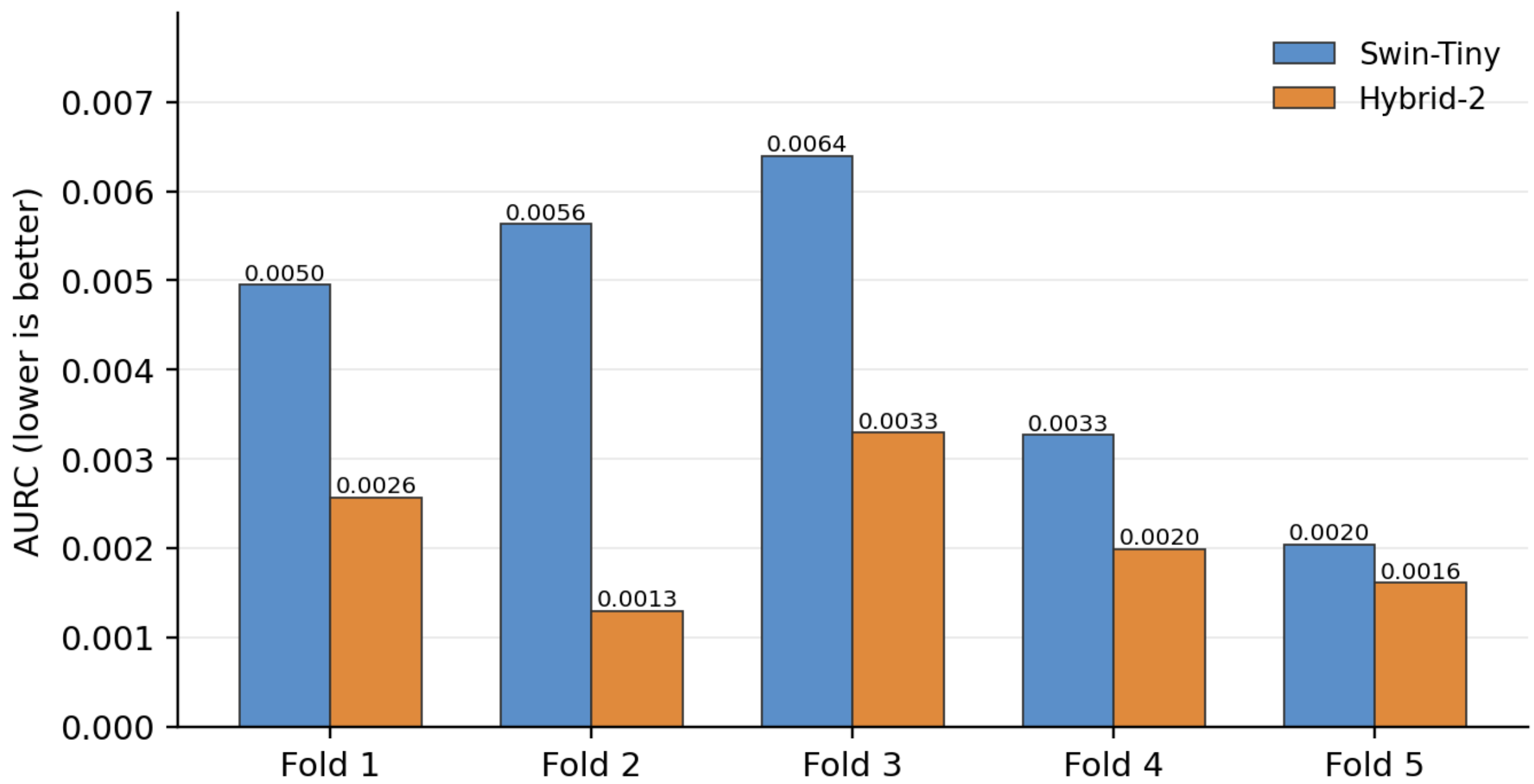

***Figure 3. Area under the risk-coverage curve (AURC, lower is better) for the best single model and the Hybrid-2 ensemble, computed independently on each of the five held-out cross-validation test folds. The ensemble achieves a lower AURC in every fold.***

Figure 4 shows the pooled risk-coverage curves. The practical consequence of the AURC difference is visible on the left of the plot: Hybrid-2 makes no errors at all until 72.8% of predictions are retained, whereas Swin-Tiny's first error appears once coverage exceeds 18.3%. In a referral setting, this is the difference between a system that can retain 72.8% of pooled cross-validation predictions before the first observed error and one that can safely auto-clear only 18.3% before its first. At full coverage the two curves converge toward their respective overall error rates, 2.73% and 3.38%, and the gap between them narrows to a margin that the accuracy comparison in Table 3 already described as non-significant.

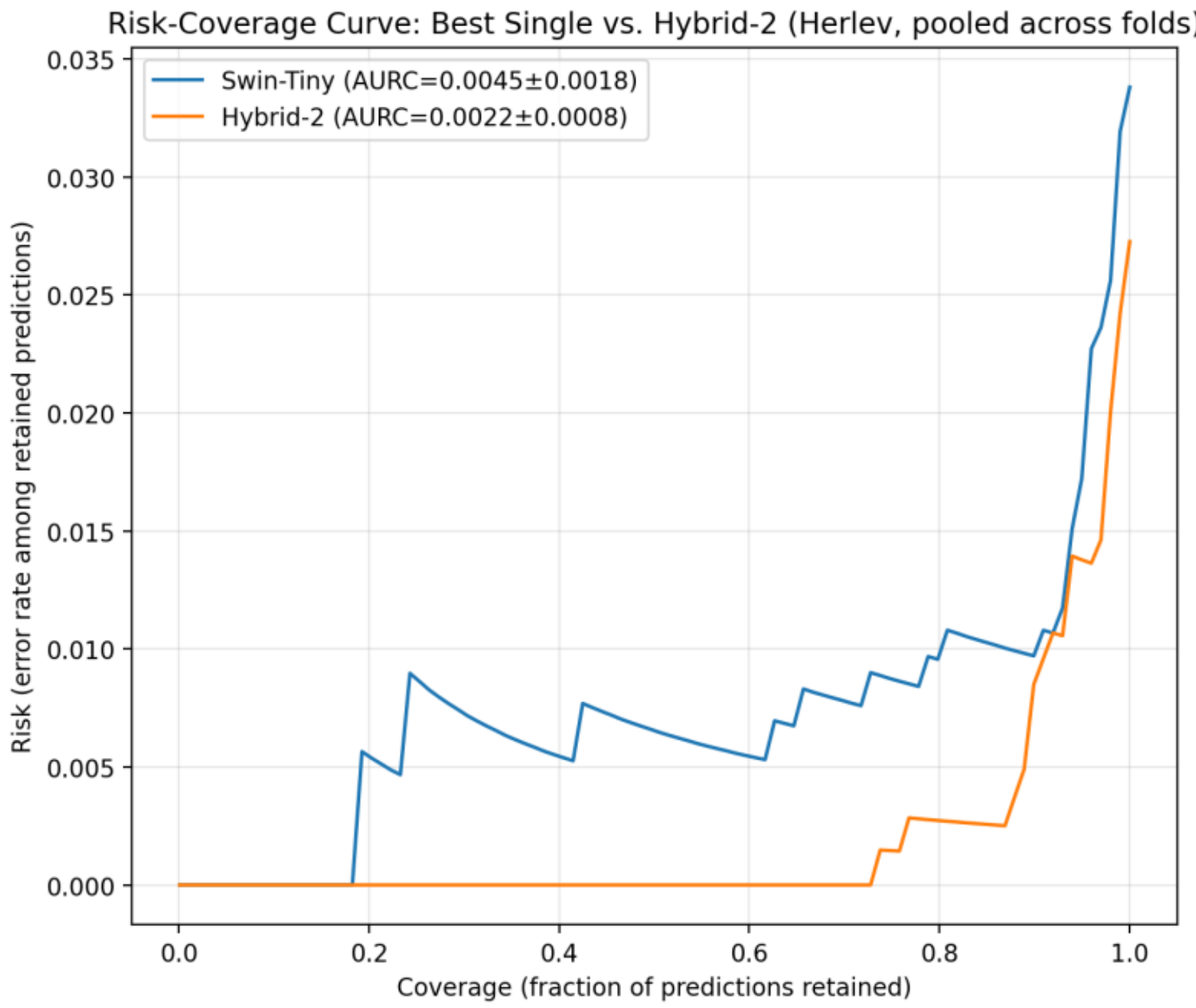

***Figure 4. Pooled risk-coverage curves for the best single model and the Hybrid-2 ensemble, obtained by concatenating the held-out test predictions of all five folds. Legend values report the per-fold mean and standard deviation of AURC; the plotted curves are pooled and their areas therefore differ from these means (see Section 3.6).***

Figure 5 reports pooled confusion matrices. Hybrid-2 makes fewer total errors than Swin-Tiny (25 vs. 31) and is markedly more specific, misclassifying 11 Normal images as Abnormal against Swin-Tiny's 21. It is, however, less sensitive: it misses 14 Abnormal images against Swin-Tiny's 10, giving sensitivities of 0.9793 and 0.9852 respectively. In cervical screening the false negative is the more costly error, since an abnormality passed as normal leaves the screening pathway entirely. The ensemble's superior aggregate error count therefore does not translate straightforwardly into superior clinical behavior, and this qualification applies to its AURC advantage as well: AURC weights all errors equally.

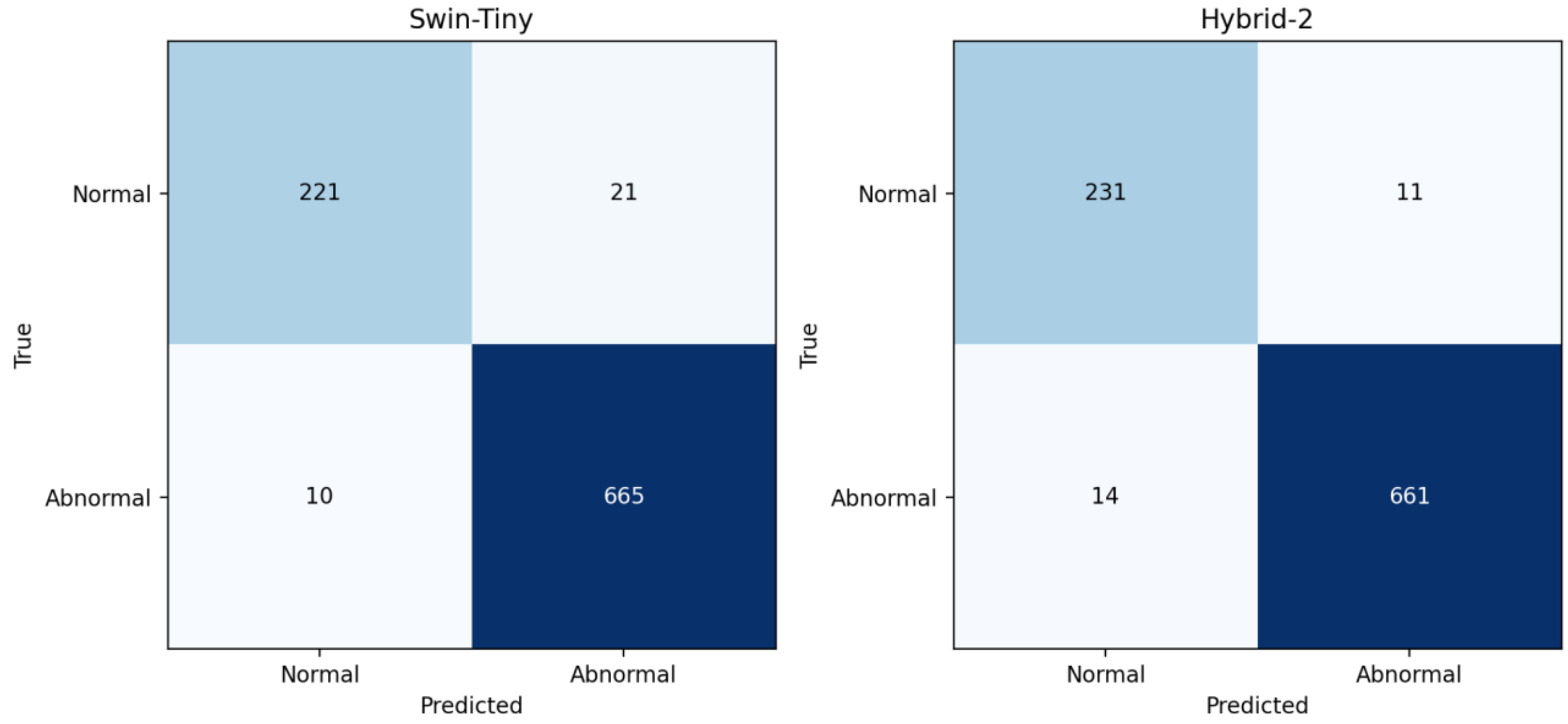

***Figure 5. Pooled confusion matrices across all 5 test folds (917 predictions) for the best single model and the Hybrid-2 ensemble, using calibrated predictions.***

*Table 5. Per-model calibration results for each individual architecture, averaged across the five cross-validation folds. T is the fitted temperature. All metrics computed on held-out test partitions only.*

| Model | T | Acc. (after) | Macro-F1 (after) | AUROC (after) | ECE (before) | ECE (after) |
|---|---|---|---|---|---|---|
| Swin-Tiny | 0.655 | 0.9662 | 0.9559 | 0.9907 | 0.0527 | 0.0247 |
| TinyViT-5M | 0.587 | 0.9640 | 0.9545 | 0.9912 | 0.0594 | 0.0303 |

Table 5 disaggregates the two individual models. Both fit temperatures below one (0.655 and 0.587), indicating that both are under-confident on this dataset rather than over-confident, a direction opposite to that typically reported for large networks on natural-image benchmarks. Neither temperature approaches the bounds of the search interval. The two models are close on every metric, which is consistent with the ensemble of the two behaving similarly to either member on accuracy while differing from both on how its confidence is distributed.

## 5. Discussion

The central finding of this study is that ensembling improved the ranking of predictions by confidence while degrading the accuracy of the confidence values themselves. Hybrid-2 halved AURC relative to the best single model, and did so in every fold, yet its post-calibration ECE was higher than that of the single model in four folds out of five. These are not contradictory results. ECE measures whether a stated confidence of p is correct p of the time; AURC measures only whether predictions with higher confidence are more often correct than predictions with lower confidence. Averaging the probabilities of two models pulls confidence values toward the middle of the range, which harms the first quantity, while suppressing the idiosyncratic high-confidence errors of either member, which helps the second.

For a referral system, confidence-based ranking directly governs the observed risk–coverage tradeoff and therefore how much workload can be retained at a given observed risk. The operating characteristic that matters is where the risk-coverage curve leaves zero, and on that measure the ensemble is not marginally better but qualitatively so: 72.8% coverage at zero pooled risk against 18.3%. Within the pooled cross-validation predictions, the ensemble deferred fewer than one-third of cases before the first observed error among retained predictions. This operating point should not be interpreted as a clinically validated deployment threshold. This is the practical argument for evaluating selective prediction directly rather than inferring it from accuracy, for which no statistically significant difference between the two configurations was detected in this study.

Two qualifications temper the conclusion, and both point the same direction. First, the ensemble is less sensitive than the single model, missing 14 abnormal images against 10. AURC treats a false negative and a false positive as equivalent, so a metric that ranks Hybrid-2 as clearly superior is silent about the error type that matters most in cervical screening. A cost-sensitive analogue of the risk-coverage curve, in which risk is defined by clinical cost rather than raw error, would be the appropriate next instrument, and this study does not provide one. Second, the reversed class balance of Herlev, in which abnormal images outnumber normal ones by nearly three to one, is an artifact of dataset construction and the opposite of a screening population. Sensitivity and specificity estimated on this distribution should not be read as estimates of screening performance.

The findings should be interpreted within the scope of this dataset, model pool, and calibration protocol. Herlev is small (917 single-cell images), and the calibration subset carved out of each fold's training portion is correspondingly small, which is reflected in the fold-to-fold variability of the fitted temperatures. The candidate pool contains only two architectures, so the ensemble is the entire pool and no model selection is exercised; whether the divergence between calibration and ranking persists for larger and more diverse ensembles is an open question that a two-model study cannot answer. Finally, the binary Normal-versus-Abnormal formulation, though conventional for Herlev, discards the low-grade versus high-grade distinction that a clinical triage system would need to make. The small candidate pool also reflects a preference for parsimonious modeling: when data are limited, a model built from a few informative features can generalize better than a larger one, as we found in predicting post-stroke functional outcomes [32]. The authors' broader work spans other machine-learning domains, including speaker identification [33] and underwater sensor networks [34].

Read alongside our recent study on liquid-based cervical cytology, in which ensembling provided no consistent reliability benefit once individual models were properly calibrated [20], the present result suggests that the value of an ensemble depends on which reliability property is being measured. Where that study asked whether ensembling improved calibration and found that it did not, this one asks whether ensembling improves selective prediction and finds that it does, on a different dataset and class formulation. The two findings are compatible, and together they caution against treating ensemble size as a single lever on a single quantity called reliability.

## 6. Conclusion

This paper evaluated selective prediction and uncertainty-aware referral for Pap smear classification on the Herlev dataset under a binary Normal-versus-Abnormal formulation, fine-tuning two lightweight transformer backbones on Herlev from ImageNet-pretrained weights and comparing the better individual model against a soft-voting ensemble of both. Temperature scaling, fit on a calibration subset held out from both training and test data, reduced expected calibration error by roughly half for both configurations without altering discrimination. No statistically significant difference was detected between the ensemble and the best single model in accuracy or macro-F1, and the ensemble was less well calibrated than the best single model, yet it halved the area under the risk-coverage curve, did so consistently in all five folds, and extended the coverage at which no errors are made from 18.3% to 72.8% of pooled predictions. Its advantage in aggregate error count was accompanied by a loss of sensitivity, the error type most costly in screening. These results indicate that the ability to rank predictions by trustworthiness and the accuracy of the confidence values themselves are distinct properties that respond differently to ensembling, and that a system intended to defer its uncertain cases should be evaluated on the former directly rather than through accuracy or calibration alone.